\documentclass[letterpaper, 10 pt, journal, twoside]{IEEEconf}
\IEEEoverridecommandlockouts
 
\usepackage{cite}
\usepackage{amsmath,amssymb,amsfonts}
\usepackage{graphicx}
\usepackage{textcomp}
\usepackage{color}
\usepackage{subfigure}
\usepackage{multirow}
\usepackage{multicol}
\usepackage{makecell}
\usepackage{bm}
\usepackage{algorithm}
\usepackage{algpseudocode}
\usepackage{hyperref}
\newcommand{\ie}{i.e.\ }

\newcommand{\Reffig}[1]{Figure~\ref{#1}}
\newcommand{\Refsec}[1]{Section~\ref{#1}}

\newcommand{\Reftab}[1]{Table~\ref{#1}}

\definecolor{darkyellow}{rgb}{1.0, 0.5,.0}

\hypersetup{hidelinks,
        colorlinks=true,
        allcolors=blue,
        pdfstartview=Fit,
        breaklinks=true}
\usepackage{graphics} 
\usepackage{epsfig} 
\usepackage{algorithm}
\usepackage{algpseudocode}
\usepackage{amsmath}
\usepackage{amssymb}
\usepackage{subfigure}
\usepackage{graphicx}
\usepackage{subfigure}
\usepackage{booktabs}
\usepackage{threeparttable}
\usepackage{multirow}
\usepackage{multicol}
\usepackage{makecell}
\usepackage{dcolumn}
\usepackage[T1]{fontenc}
\usepackage{diagbox}

\begin{document}
\title{
        LOG-LIO2: A LiDAR-Inertial Odometry with Efficient Uncertainty Analysis
}

\author{Kai Huang$^{1}$, Junqiao Zhao$^{*,2, 3}$, Jiaye Lin$^{2, 3}$, Zhongyang Zhu$^{2, 3}$, Shuangfu Song$^{1}$, Chen Ye$^{2}$, Tiantian Feng$^{1}$ 
\thanks{$^{1}$Kai Huang, Shuangfu Song and Tiantian Feng are with the School of Surveying and Geo-Informatics, Tongji University, Shanghai, China
{\tt\footnotesize (e-mail: huangkai@tongji.edu.cn; songshuangfu@tongji.edu.cn; fengtiantian@tongji.edu.cn).}}
\thanks{$^{2}$Junqiao Zhao, Jiaye Lin, Zhongyang Zhu and Chen Ye are with Department of Computer Science and Technology, School of Electronics and Information Engineering, Tongji University, Shanghai, China, and the MOE Key Lab of Embedded System and Service Computing, Tongji University, Shanghai, China
{\tt\footnotesize (e-mail: zhaojunqiao@tongji.edu.cn; 2233057@tongji.edu.cn; 2310920@tongji.edu.cn; yechen@tongji.edu.cn).}}
\thanks{$^{3}$Institute of Intelligent Vehicles, Tongji University, Shanghai, China.}
\thanks{Digital Object Identifier (DOI): see top of this page.}
}


\maketitle

\begin{abstract}
Uncertainty in LiDAR measurements, stemming from factors such as range sensing, 
is crucial for LIO (LiDAR-Inertial Odometry) systems as it affects the accurate weighting in the loss function. 
While recent LIO systems address uncertainty related to range sensing, the impact of incident angle on uncertainty is often overlooked by the community. 
Moreover, the existing uncertainty propagation methods suffer from computational inefficiency.
This paper proposes a comprehensive point uncertainty model that accounts for both the uncertainties from LiDAR measurements and surface characteristics, along with an efficient local uncertainty analytical method for LiDAR-based state estimation problem.
We employ a projection operator that separates the uncertainty into the ray direction and its orthogonal plane.
Then, we derive incremental Jacobian matrices of eigenvalues and eigenvectors w.r.t. points, which enables a fast approximation of uncertainty propagation. 
This approach eliminates the requirement for redundant traversal of points, significantly reducing the time complexity of uncertainty propagation from $\mathcal{O} (n)$  to $\mathcal{O} (1)$ when a new point is added. 
Simulations and experiments on public datasets are conducted to validate the accuracy and efficiency of our formulations.
The proposed methods have been integrated into a LIO system, which is available at \href{https://github.com/tiev-tongji/LOG-LIO2}{https://github.com/tiev-tongji/LOG-LIO2}.
\end{abstract}




\section{INTRODUCTION}
\label{sec:introduction} 


Uncertainty stemming from sensor limitations, measurement noises, and unpredictable physical environments plays a crucial role in robotic state estimation, as it affects the accurate weighting of distance metrics in the loss function \cite{barfoot2024state}.
One common approach is to model the measurement uncertainty using zero-mean Gaussian noise with respect to point coordinates and integrate these noises into the state estimation.
However, such uncertainty modeling of LiDAR measurements lacks sufficient accuracy.

To address this issue, \cite{yuan2021pixel} develops a point uncertainty model that incorporates both range and bearing, while \cite{jiang2022lidar} further includes surface roughness. 
These uncertainties are then propagated to the geometric elements such as planes, which are utilized for weighting the point-to-element distance in scan-to-map registration, thereby enhancing the accuracy and reliability of the system \cite{yuan2021pixel, liu2021balm, yuan2022efficient}.

However, range uncertainty is also influenced by the incident angle \cite{tasdizen2003cramer, bae2008closed}, a factor often overlooked by existing studies.
Additionally, continuous propagation of uncertainty from numerous points, which exhibits linear time complexity $\mathcal{O} (n)$, can be computationally expensive, particularly for dense point cloud and real-time applications.
Therefore, an efficient uncertainty propagation method is highly desirable.

In this paper, we present a comprehensive point uncertainty model that incorporates all relevant factors affecting LiDAR measurements, including range, bearing, incident angle, and surface roughness. 
This model leverages the projection operator \cite{bae2008closed} to separate the uncertainty into the ray direction and its orthogonal plane. 

To accelerate the propagation of uncertainty from points to the geometric elements, 
we derive the incremental Jacobian matrices for eigenvalues and eigenvectors corresponding to specified points from Welford's formulation \cite{welford1962note}.
The parametric uncertainty of the geometric elements is then updated incrementally by fast approximations.

We validate the efficiency and accuracy of the fast uncertainty approximation through simulation experiments. 
Subsequently, we integrate these methods into a LIO system named LOG-LIO2. 
By comparing the performance of LOG-LIO2 with state-of-the-art LIO systems, \ie VoxelMap \cite{yuan2022efficient}, LOG-LIO \cite{huang2023log}, and FastLIO2 \cite{xu2022fast}, we demonstrate the performance improvement achieved by the proposed techniques. 

The main contributions of this work are as follows:
\begin{itemize}
        \item
                A comprehensive point uncertainty model with a fast calculation method,
                incorporating the uncertainty of range and bearing measurements from LiDAR,
                as well as incident angle and roughness concerning the target surface. 
        \item
                A fast approximation of uncertainty propagation from points to eigenvalues and eigenvectors by leveraging the incremental Jacobian matrices, 
                reducing the time complexity from $\mathcal{O} (n)$  to  $\mathcal{O} (1)$ by eliminating the need for repetitive calculations when a new point is added.
        \item
              We demonstrate the efficiency and accuracy of our methods from both simulation and public datasets.
              The methods proposed in this paper are incorporated into a LIO system which is available at \href{https://github.com/tiev-tongji/LOG-LIO2}{https://github.com/tiev-tongji/LOG-LIO2}.
\end{itemize}

\section[sec:related_work]{RELATED WORKS} 
\label{sec:related_work}

The integration of IMU has significantly improved the practical robustness of LIO systems \cite{xu2022fast}.
However, a critical limitation persists in these systems: the uncertainty of each point is treated homogeneously, neglecting potential spatial and environmental variations that can significantly impact accuracy.

\cite{yuan2021pixel} investigates the physical measuring principles of LiDAR and derives the uncertainty model for each laser beam encompassing range and bearing uncertainties.
Specifically, the bearing uncertainty is modeled as a $\mathbb{S}^2$ perturbation \cite{he2021kalman} in the tangent plane of the ray.
However, the calculation of $\mathbb{S}^2$ perturbation requires the construction of two bases in the tangent plane, leading to increased computational cost.

\cite{liu2021balm} introduces a novel adaptive voxelization method that iteratively divides voxels until the point distribution matches the line or plane geometry.
The paper derives the Jacobian and Hessian matrices of eigenvalues and eigenvectors w.r.t. the points within the voxel, leveraging these matrices for LiDAR bundle adjustment. 

Building on these prior works, VoxelMap \cite{yuan2022efficient} integrates the point uncertainty model and adaptive voxelization technique into a LiDAR-based odometry system. 
The uncertainty associated with each point is then propagated to the plane parameters to apply weights to the point-to-plane distance in the state estimation.
However, the traversing of all points during propagation reduces the real-time performance of the system.

In addition to inherent laser range and bearing uncertainties, the geometrical properties of the target surface also contribute significantly to observed point uncertainty.
\cite{tasdizen2003cramer} and \cite{bae2008closed} analyze the geometrical relationship between the laser beam and the surface normal, incorporating the uncertainty arising from the incident angle into the ray direction.
While they obtain a closed-form approximation of this uncertainty, it is overlooked by the SLAM community.

Distinct from prior works, \cite{jiang2022lidar} introduces a principled uncertainty model for LiDAR scan-to-map registration.
This model incorporates the roughness of the measured uneven planes, thereby enhancing robustness but also introducing complexity in estimation. 

Having the above, our focus is on developing a comprehensive point uncertainty model and deriving efficient uncertainty propagation methods, aiming to strike a balance between accuracy and computational efficiency.

\section{PRELIMINARY}
\label{sec:preliminary}
In this paper, we assume the transformation between LiDAR and IMU is calibrated in advance.
We denote the statistics of point cloud $\mathcal{P}= \{\mathbf{p}_i,i=1,...,k \}$ by $(\cdot )_k$, where the subscript $k$ indicates the point cloud with $k$ points.
For simplicity, the parentheses "()" may be omitted when there is no ambiguity. 
The sum of the outer products of all points deviating from the center $\mathbf{m}$ is denoted by $\mathbf{S}$, which is normalized to obtain the covariance matrix, $\mathbf{A}$.
\begin{equation}
        \mathbf{m}_k=\frac{1}{k} \sum_{i=1}^{k}\mathbf{p}_{i}; \mathbf{A}_k = \frac{1}{k} \mathbf{S}_k =\frac{1}{k} \sum_{i=1}^{k}(\mathbf{p}_{i} - \mathbf{m}_k)(\mathbf{p}_{i} -  \mathbf{m}_k)^{T}.
        \label{eq_cov}
\end{equation}

For the symmetric matrix $\mathbf{A}$, eigenvalue decomposition is performed as follows:
\begin{equation}
        \mathbf{A} = \mathbf{V} \mathbf{\Lambda}\mathbf{V}^T, 
        \mathbf{V} = \begin{bmatrix}
                \mathbf{v}_1 &  \mathbf{v}_2 &  \mathbf{v}_3
               \end{bmatrix}
        \label{eq_pca},
        \mathbf{\Lambda}=diag(\lambda_1, \lambda_2, \lambda_3)
\end{equation}
where $\mathbf{v}_i$ is the eigenvector corresponding to the eigenvalue $\lambda_i$ with $\lambda_1 < \lambda_2 <\lambda_3$.

\section{Point Uncertainty Model}
\label{sec:uncer_model}
The point-wise uncertainty model quantifies the reliability of each input point, allowing SLAM systems to handle real-world complexities more effectively. 
Modeling the point uncertainty derived from the range and bearing measurements as \cite{yuan2021pixel, liu2021balm, yuan2022efficient} is natural since the coordinate is obtained from the relative position w.r.t. the LiDAR.
However, the geometric properties of the target surface, \ie incident angle and roughness, also contribute to the uncertainty of the LiDAR measurements \cite{jiang2022lidar, tasdizen2003cramer, bae2008closed}. 
Our uncertainty model takes into account both the properties of the sensor itself and the observed geometry.

\begin{figure}[!ht]
        \centering
        \subfigure[point uncertainty model]
        {
                \begin{minipage}[b]{0.23\textwidth}
                        \includegraphics[width=1\textwidth]{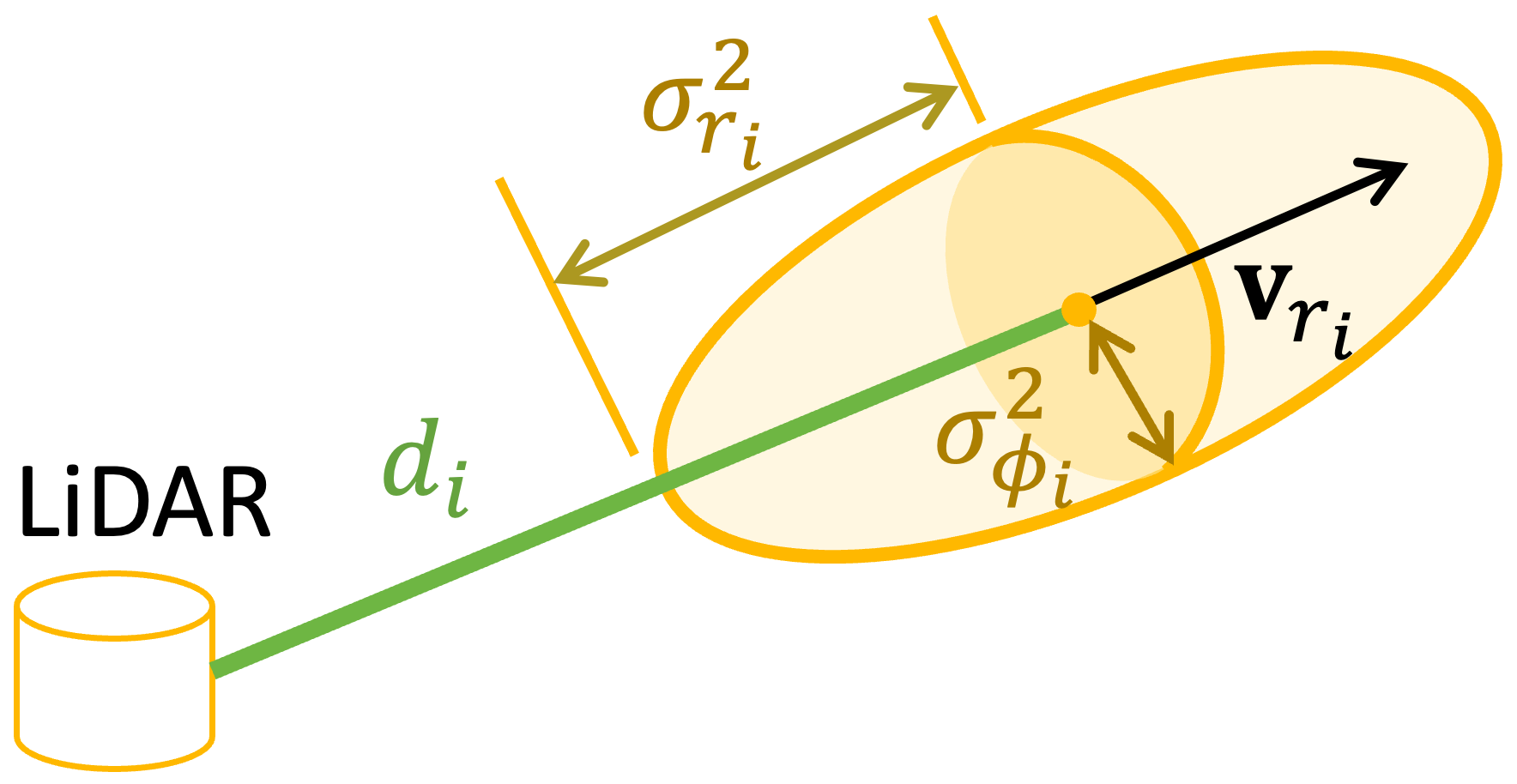}
                \end{minipage}
                \label{fig_p_uncer}

        }
        \subfigure[incident angle and roughness]
        {
                \begin{minipage}[b]{0.22\textwidth}
                        \includegraphics[width=1\textwidth]{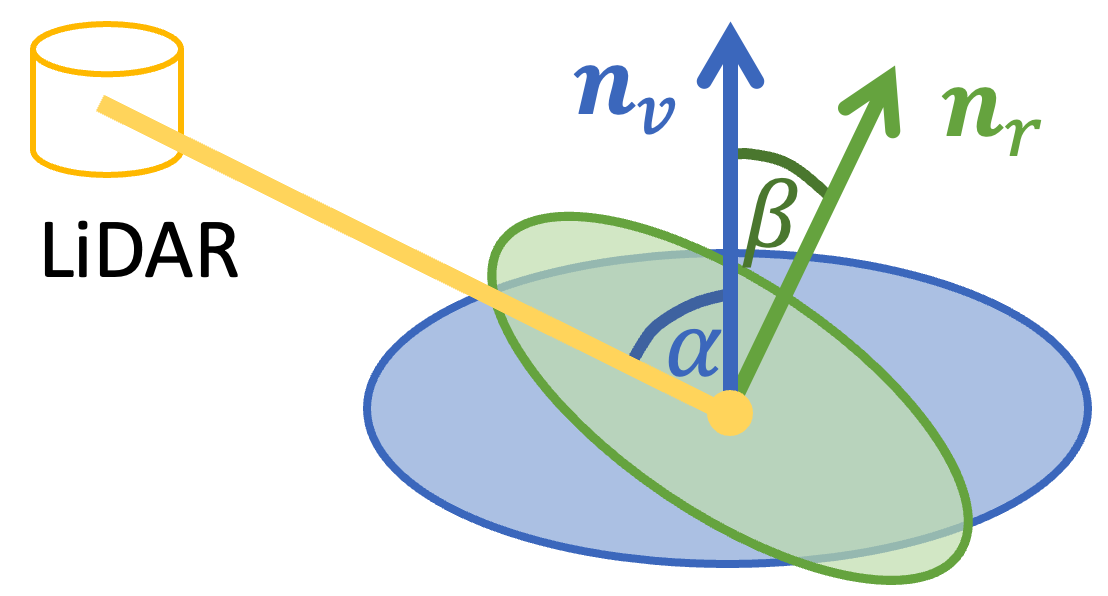}
                \end{minipage}
                \label{fig_inc_rou}
        }
        \caption{Illustrations of geometric relationship for the point uncertainty model.
                }
\end{figure}

We assume the point-wise uncertainty follows a Gaussian distribution.
Inspired by \cite{bae2008closed} and as shown in \Reffig{fig_p_uncer}, we denote the magnitude of uncertainty along the ray $\mathbf{v}_{r_i}$ direction as $\sigma_{r_i}^2$ 
while the magnitude of uncertainty on the plane orthogonal to $\mathbf{v}_{r_i}$ is denoted as $\sigma_{\phi  _i }^2$.
Note that the orientation of $\sigma_{\phi _i}^2$ is not specified in order to simplify the calculations, as further explained in \Refsec{sec:fast_uncer}.

\subsection{Uncertainty Factors}
\label{sec:unc_f}
\subsubsection{Range and bearing}
\label{sec:lid_mea}
Similar to \cite{yuan2021pixel, yuan2022efficient}, but with a different calculation approach,
we assume that the magnitude of the range and bearing uncertainty to each laser beam are both constant factors, $\sigma_{d}^2$ and $\sigma_{\omega}^2$, respectively.
In a 3D space, the $\sigma_{d}^2$ extends in the $\mathbf{v}_{r_i}$ direction, while $\sigma_{\omega}^2$ is isotropic in the orthogonal plane of the $\mathbf{v}_{r_i}$.

\subsubsection{Incident Angle}
\label{sec:incident}
As illustrated in \Reffig{fig_inc_rou}, the incident angle $\alpha $ between the laser beam and the surface normal plays a pivotal role in the range sensing \cite{tasdizen2003cramer, bae2008closed}.
As $\alpha$ increases from 0 to $\pi /2$, the range uncertainty also rises.
We adopt the uncertainty model caused by the incident angle $\sigma_{in_i}$ along the ray direction from \cite{tasdizen2003cramer, bae2008closed}:
\begin{equation}
        \begin{aligned}
                \sigma_{in_i} &= d_i \sigma_{\omega}\tan\alpha_i
        \end{aligned}
        \label{eq_incident}
\end{equation}
where $d_i$ is the distance from $\mathbf{p}_i$ to the LiDAR.

\subsubsection{Roughness}
\label{sec:roughness}
We account for the impact of surface roughness on the point in a simpler and faster manner compared to \cite{jiang2022lidar}.
To achieve this, we estimate the normal of the target surface by analyzing two different neighborhood scales. 
As shown in \Reffig{fig_inc_rou}, the angle $\beta$ formed between these two normals serves as a metric for quantifying the uncertainty introduced by the roughness:
\begin{equation}
        \sigma_o = \eta \sin\beta, \beta = \arccos(\mathbf{n}_r,\mathbf{n}_v)
        \label{eq_roughness}
\end{equation}
where $\eta$ is a preset value,
$\mathbf{n}_r$ and $\mathbf{n}_v$ are normals estimated under different neighborhood scales.

\subsection{Fast Uncertainty Calculation}
\label{sec:fast_uncer}
Assuming the roughness is isotropic in 3D space,
we integrate the aforementioned three factors to compute the covariance of the point $\mathbf{p}_i$ as follows:
\begin{equation}
        \mathbf{A}_{\mathbf p_i} = \sigma_{r_i}^2 \mathbf{v}_{r_i} \mathbf{v}_{r_i}^T + \sigma_{\phi _i}^2(\mathbf{I}_{3\times3} -\mathbf{v}_{r_i} \mathbf{v}_{r_i}^T) + \sigma_o^2\mathbf{I}_{3\times3}  
        \label{eq_sigmapl}
\end{equation}
where $\sigma_{r_i}^2 = \sigma_{d}^2 + \sigma_{in_i}^2$, $\sigma_{\phi _i} = d_i\sigma_{\omega}$, and $\mathbf{v}_{r_i}$ is the unit vector in ray direction.
The first term  captures the uncertainty arising from the range measurement and incident angle.
We use the projection operator $(\mathbf{I} -\mathbf{v}_r \mathbf{v}_r^T)$ from \cite{bae2008closed} to isotropically project bearing uncertainty onto the 2D plane orthogonal to $\mathbf{v}_{r_i}$.
Note that considering only the uncertainty of range and bearing, our method aligns with \cite{yuan2021pixel, liu2021balm, yuan2022efficient}, which is proved in Appendix \ref{sec:p_po}. 
We present a more comprehensive point uncertainty model accompanied by  an efficient method that offers greater geometric interpretability.

\section{LUFA: Local Uncertainty Fast Approximation}
\label{sec:lufa}

LiDAR SLAM systems maintain statistics of localized map regions, such as mean position, eigenvalues, and eigenvectors, which are crucial for state estimation.
Deriving Jacobian matrices for these statistics is vital for uncertainty propagation using linear Gaussian principles \cite{barfoot2024state}.
As outlined  in BALM \cite{liu2021balm} and VoxelMap \cite{yuan2022efficient}
, computing these Jacobian matrices involves traversing all points in 
$\mathcal{P}$, resulting in $\mathcal{O} (n)$  time complexity, which can impact real-time performance.

To relieve this computation burden, we propose an incremental update approach for computing these statistics in $\mathcal{O} (1)$ time complexity when adding a new point.
This ensures real-time performance, ideal for SLAM applications.

\subsection{Incremental Center and Covariance}
\label{sec:iCov}
Using Welford's formula \cite{welford1962note}, we incrementally update the center $\mathbf{m}$ and covariance $\mathbf{S}$ of $\mathcal{P}$ as follows:
\begin{equation}
        \mathbf{m}_k = \frac{(k-1)}{k}\mathbf{m}_{k-1}+\frac{1}{k}\mathbf{p}_k
        \label{eq_mk}
\end{equation}
\begin{equation}
        \mathbf{S}_k = \mathbf{S}_{k-1} + \frac{k-1}{k}(\mathbf{p}_k - \mathbf{m}_{k-1})(\mathbf{p}_k - \mathbf{m}_{k-1})^T
        \label{eq_sk}
\end{equation}
where $\mathbf{m}_k$ and $\mathbf{S}_k$ are updated from $\mathbf{m}_{k-1}$ and $\mathbf{S}_{k-1}$, $\mathbf{p}_k$ is the newly added point.
Define $ \mathbf{D} = (\mathbf{p}_k - \mathbf{m}_{k-1})(\mathbf{p}_k - \mathbf{m}_{k-1})^T$ and combine (\ref{eq_cov}),
the normalized covariance $\mathbf{A}$ can be updated incrementally as follows:
\begin{equation}
        \begin{aligned}
                \mathbf{A}_k &= \frac{k-1}{k}\mathbf{A}_{k-1} + \frac{k-1}{k^2}\mathbf{D}
        \end{aligned}
        \label{eq_AkAk1}
\end{equation}

The updated covariance $\mathbf{A}_k$ decomposes into two terms: 
the former scales the existing point cloud's covariance, while the latter captures the contribution from the new point $\mathbf{p}_k$.

\subsection{Incremental Jacobian of Eigenvalues}
\label{sec:i_jevalue}
Deducing from \eqref{eq_AkAk1},
the partial derivative of the $j$-th eigenvalue $\lambda_j$ w.r.t. the point $\mathbf{p}_i, i < k$, can be updated incrementally as follows:
\begin{equation}
        \begin{aligned}
                \frac{\partial \lambda_{j,k}}{\partial \mathbf{p}_i } &= \frac{k-1}{k}\frac{\partial \lambda_{j,k-1}}{\partial \mathbf{p}_i} \\
                &-\frac{d_u}{k^2 \cos\theta_j}(\cos\varphi_{j,k-1}\mathbf{v}_{j,k}^T + \cos\varphi_{j,k}\mathbf{v}_{j,k-1}^T)   
        \end{aligned}
        \label{eq_dlk_dpj}
\end{equation}
where $d_u $ is the distance between $\mathbf{p}_k$ and $\mathbf{m}_{k-1}$, 
$\theta_j$ is the angle between $\mathbf{v}_{j,k}$ and $\mathbf{v}_{j,k-1}$.
Define $\mathbf{v}_u = (\mathbf{p}_k - \mathbf{m}_{k-1})$, 
then $\varphi_{j,k}$ and $\varphi_{j,k-1}$ are the angles from the direction of $\mathbf{v}_u$ to the $j$-th eigenvector $\mathbf{v}_{j,k}$ derived from $k$ points and the $j$-th eigenvector $\mathbf{v}_{j,k-1}$ derived from $k-1$ points, respectively.
The proof of (\ref{eq_dlk_dpj}) is provided in Appendix \ref{sec:p_ije}.

\begin{figure}[!ht]
        \centering
        {
                \begin{minipage}[b]{0.46\textwidth}
                        \includegraphics[width=1\textwidth]{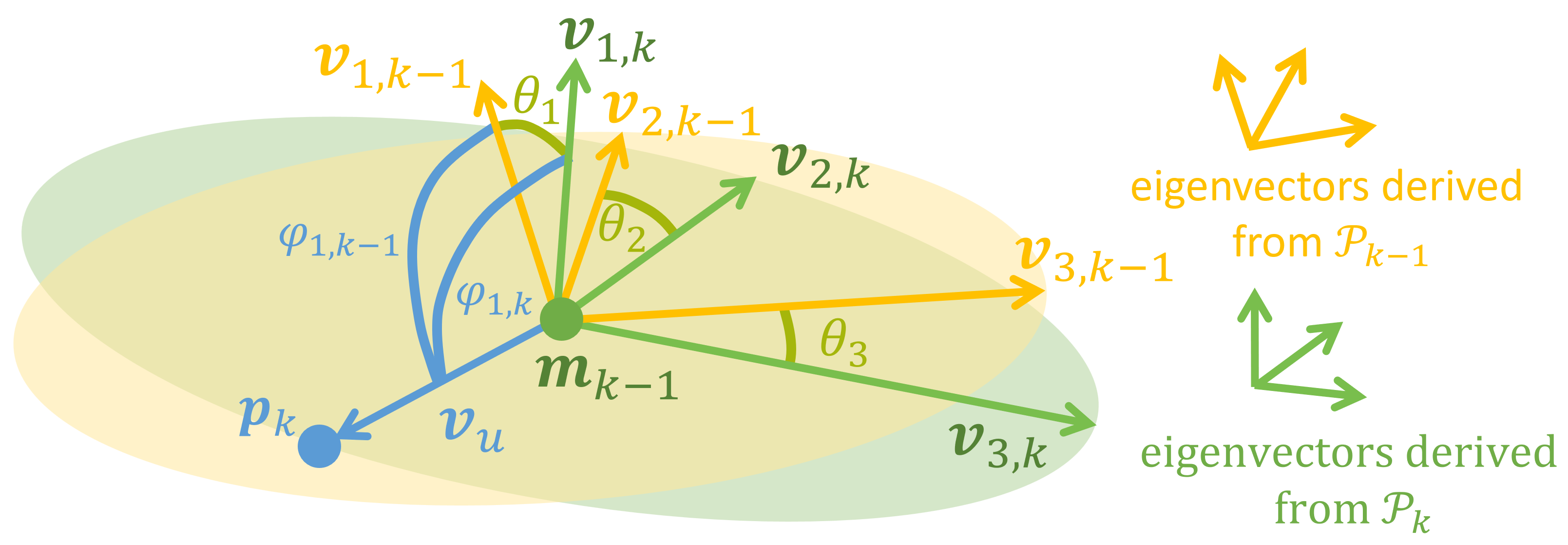}
                \end{minipage}
        }
        \caption{
                An illustration of geometric relationship for the $\mathbf{m}_{k-1}$ local frame.
                }
        \label{fig_m_local}
\end{figure}

\Reffig{fig_m_local} visually depicts the geometric relationships involved in \eqref{eq_dlk_dpj}. 
To clarify the geometric meaning, let's consider $\mathbf{v}_1$, the normal of a plane, as an example, which determines the direction for calculating the point-to-plane distance.
Suppose we have sampled a sufficiently large number of points from the plane and $\mathbf{p}_k$ lies on it, then, $\theta_1 \approx 0$ and $\varphi_{1,k-1} = \varphi_{1,k} \approx \pi / 2$.
Consequently, the second term of $ {\partial \lambda_{1,k}}/{\partial \mathbf{p}_i }$ is approximately equal to \textbf{0}.

Furthermore, the second term (incremental terms) of (\ref{eq_dlk_dpj}) are identical for the first $k-1$ points. 
This characteristic allows for their efficient computation without the need to traverse $\mathcal{P}$.
We take the magnitude of the second term of \eqref{eq_dlk_dpj} to determine the updated form of the Jacobian matrix of the eigenvalues.
If the magnitude is smaller than a threshold, we approximately update the Jacobian of eigenvalues as follows:
\begin{equation}
        \mathbf{J}_{\lambda_j,\mathbf{p}_i,k} = \frac{\partial \lambda_{j,k}}{\partial \mathbf{p}_i} \approx \frac{k-1}{k}\frac{\partial \lambda_{j,k-1}}{\partial \mathbf{p}_i}, i < k.
        \label{eq_jlkk1}
\end{equation}
Otherwise, the Jacobian is updated as in BALM \cite{liu2021balm}:
\begin{equation}
        \begin{aligned}
                \frac{\partial \lambda_{j,k}}{\partial \mathbf{p}_i} 
                &= \frac{2}{k}(\mathbf{p}_i - \mathbf{m}_k)^T\mathbf{v}_{j,k}\mathbf{v}_{j,k}^T.
        \end{aligned}
        \label{eq_dlk_balm}
\end{equation}

Considering that $\mathbf{p}_k$ is independent of $\mathbf{A}_{k-1}$, the Jacobian matrix of eigenvalue w.r.t. $\mathbf{p}_k$ can be simplified as:
\begin{equation}
        \begin{aligned}
                \frac{\partial \lambda_{j,k}}{\partial \mathbf{p}_k} 
                &= \frac{d_u(k-1)}{k^2 \cos\theta_j}(\cos\varphi_{j,k-1}\mathbf{v}_{j,k}^T + \cos\varphi_{j,k}\mathbf{v}_{j,k-1}^T),
        \end{aligned}
        \label{eq_dlk}
\end{equation}
which is proved in Appendix \ref{sec:p_ije}.

\subsection{Incremental Jacobian of Eigenvectors}
\label{sec:i_jevector}
According to BALM \cite{liu2021balm}, the Jacobian matrix of the eigenvector $\mathbf{v}_j$ w.r.t. $\mathbf{p}_i$ can be represented as
$\mathbf{J}_{\mathbf{v}_j,\mathbf{p}_i} = {\partial\mathbf{v}_j} / {\partial \mathbf{p}_i} = \mathbf{V} \mathbf{C}$, 
where $\mathbf{V}$ is the eigenmatrix as shown in \eqref{eq_pca}.
For the elements in $m$-th row and $n$-th column in $\mathbf{C}$:
\begin{equation}
        \begin{aligned}
                \mathbf{C}^{\mathbf{p}_i}_{m,n} &=\left\{\begin{aligned}
                 \frac{(\mathbf{p}_i-\mathbf{m}_k)^T}{k(\lambda_n-\lambda_m)} (\mathbf{v}_m \mathbf{v}_n^T + \mathbf{v}_n \mathbf{v}_m^T), m\neq n \\
                \mathbf{0}  \qquad\qquad\qquad, m=n.
                \end{aligned}\right .
        \end{aligned}
        \label{eq_cvv}
\end{equation}

And \eqref{eq_cvv} can be further simplified as
$\mathbf{C}_{m,n}^{\mathbf{p}_i} = \begin{bmatrix}
        \mathbf{C}^{x_i}_{m,n} & \mathbf{C}^{y_i}_{m,n} & \mathbf{C}^{z_i}_{m,n}
        \end{bmatrix} \in \mathbb{R}^{1 \times 3},  m, n \in \{ 1,2,3 \}$,
where the subscripts $x_i, y_i$, and $z_i$ are elements of $\mathbf{p}_i$.
Since diagonal elements in $\mathbf{C}$ are equal to 0,
the rest of this section only specifies the form of the off-diagonal elements.

Analogous to the Jacobian matrix of the eigenvalues, the above matrix $\mathbf{C}$ w.r.t $\mathbf{p}_i, i < k,$ can be incrementally updated,
which is also divided into scale and increment parts:
\begin{equation}
        \begin{aligned}
                (\mathbf{C}^{\mathbf{p}_i}_{m,n})_k &=  \begin{bmatrix}
                        (\mathbf{C}^{x_i}_{m,n})_k & (\mathbf{C}^{y_i}_{m,n})_{k} &( \mathbf{C}^{z_i}_{m,n})_{k}
                       \end{bmatrix}_{1\times3} \\
                &={\mathbf{W}^{\mathbf{p}_i}_{m,n}}
                (\mathbf{C}^{\mathbf{p}_i}_{m,n})_{k-1}  \\
                &-\frac{d_u} {k^2(\lambda_{n} - \lambda_{m})_k}
                (\cos\varphi_{m}\mathbf{v}_{n}^T + \cos\varphi_{n}\mathbf{v}_{m}^T)_k      
        \end{aligned}
        \label{eq_ckck1}
\end{equation}
where $\varphi_{m}$ and $\varphi_{n}$ are angles  from $\mathbf{v}_u$ to the $m$-th eigenvector $\mathbf{v}_{m}$ and the $n$-th eigenvector $\mathbf{v}_{n}$, respectively.
We specify the form of $\mathbf{W}^{\mathbf{p}_i}_{m,n}$ and prove \eqref{eq_ckck1} in Appendix \ref{sec:p_ijevectors}.

With the assumption that a sufficiently large number of points are already sampled from the surface, the second term of \eqref{eq_ckck1} is approximately equal to \textbf{0}. 
Take $\mathbf{v}_1$, \ie{normal $\mathbf{n}$}, of the surface as an example, 
the partial derivative of the updated $\mathbf{n}$ w.r.t. $\mathbf{p}_i$ can be approximately updated as follows:
\begin{equation}
        \begin{aligned}
                \mathbf{J}_{\mathbf{n},\mathbf{p}_i,k} = \mathbf{V}_k (\mathbf{C}_{\mathbf{n},\mathbf{p}_i})_k &\approx  \mathbf{V}_k \mathbf{W} (\mathbf{C}_{\mathbf{n},\mathbf{p}_i})_{k-1} \\
                &= \mathbf{V}_k \mathbf{W} \mathbf{V}_{k-1}^T \mathbf{V}_{k-1} (\mathbf{C}_{\mathbf{n},\mathbf{p}_i})_{k-1}  \\
                &= \overbrace{\mathbf{V}_k \mathbf{W} \mathbf{V}_{k-1}^T}^{{\mathbf{Q}}} \mathbf{J}_{\mathbf{n},\mathbf{p}_i,k-1}        
        \end{aligned}
        \label{eq_jnkjnki}
\end{equation}
where we omit some of the subscripts to simplify the representation.

Considering that $\mathbf{p}_k$ is independent of $\mathbf{A}_{k-1}$, the matrix $\mathbf{C}$ w.r.t. $\mathbf{p}_k$ can be simplified as: 
\begin{equation}
        \begin{aligned}
                (\mathbf{C}&^{p_k}_{m,n})_k = \begin{bmatrix}
                        (\mathbf{C}^{x_k}_{m,n})_k &( \mathbf{C}^{y_k}_{m,n})_k & (\mathbf{C}^{z_k}_{m,n})_k
                       \end{bmatrix}_{1\times3} \\
                &=\frac{d_u(k-1)}{k^2 (\lambda_{n} - \lambda_{m})_k} (\cos\varphi_{m} \mathbf{v}_{n}^T + \cos\varphi_{n}\mathbf{v}_{m}^T)_k .
        \end{aligned}  
        \label{eq_dbdqjk}
\end{equation}
We prove \eqref{eq_dbdqjk} in Appendix \ref{sec:p_ijevectors} and define $\mathbf{J}_{\mathbf{v}_j,\mathbf{p}_i,k} = {\partial \mathbf{v}_{j,k}}/{\partial \mathbf{p}_i}$  to simplify the representation.

\subsection{Incremental Update of Covariance}
\label{sec:i_uc}
Using the above incremental Jacobian matrix of the eigenvalues (\ref{eq_jlkk1}) and the eigenvectors (\ref{eq_jnkjnki}), their covariance can be updated in an approximate form.

We simply divide the covariance of the eigenvalues corresponding to specified points into scale and increment parts as follows:
\begin{equation}
        \begin{aligned}
                \mathbf{A}_{\lambda_j,k} 
                        &= \sum_{i}^{k-1} \mathbf{J}_{\lambda_j,\mathbf{p}_i,k} \mathbf{A}_{\mathbf{p}_i} \mathbf{J}_{\lambda_j,\mathbf{p}_i,k}^T + \mathbf{J}_{\lambda_j,\mathbf{p}_k,k} \mathbf{A}_{\mathbf{p}_k} \mathbf{J}_{\lambda_j,\mathbf{p}_k,k}^T \\
                        &= \frac{(k-1)^2}{k^2} \mathbf{A}_{\lambda_j,k-1} +  \mathbf{J}_{\lambda_j,\mathbf{p}_k,k} \mathbf{A}_{\mathbf{p}_k} \mathbf{J}_{\lambda_j,\mathbf{p}_k,k}^T 
        \end{aligned}
\end{equation}

Similarly, the covariance of the eigenvector can be updated as follows:
\begin{equation}
        \begin{aligned}
                \mathbf{A}_{\mathbf{v}_j,k} &= \mathbf{Q} \mathbf{A}_{\mathbf{v}_j,k-1} \mathbf{Q}^T + \mathbf{J}_{\mathbf{v}_j,\mathbf{p}_k,k} \mathbf{A}_{\mathbf{p}_k} \mathbf{J}_{\mathbf{v}_j,\mathbf{p}_k,k}^T \\
        \end{aligned}
        \label{eq_anqkk1}
\end{equation}
where we omit some of the subscripts to simplify the representation.

This fast covariance approximation method can be seamlessly integrated with other statistical metrics that share similar scaling properties. 
Derived from \eqref{eq_mk}, we get the Jacobian matrix of the center $\mathbf{m}_k$ w.r.t $\mathbf{p}_i$ as $\mathbf{J}_{\mathbf{m}, \mathbf{p}_i,k} = \frac{\partial \mathbf{m}_k}{\partial \mathbf{p}_i} = \frac{1}{k}\mathbf{I}_{\text{3x3}}$ easily.
Furthermore, $\mathbf{J}_{\mathbf{m},\mathbf{p}_i}$ can be incrementally updated as follows:
\begin{equation}
        \begin{aligned}
                \mathbf{J}_{\mathbf{m},\mathbf{p}_i,k} = \frac{k-1}{k}\mathbf{J}_{\mathbf{m},\mathbf{p}_i,k-1}.
        \end{aligned}
        \label{eq_jqkk1}
\end{equation}
Combining \eqref{eq_anqkk1} and \eqref{eq_jqkk1}, the covariance of 
the normal $\mathbf{n}$ and the center  $\mathbf{m}$ can be fast approximated as follows:
\begin{equation}
        \begin{aligned}
                \mathbf{A}_{\mathbf{n,m},k} =& \begin{bmatrix}
                        \mathbf{Q} \mathbf{A}_{\mathbf{n},k-1} \mathbf{Q}^T & \frac{k-1}{k}\mathbf{Q}\mathbf{A}_{\mathbf{n}\mathbf{m},k-1} \\
                        \frac{k-1}{k}\mathbf{A}_{\mathbf{m}\mathbf{n}, k-1}  \mathbf{Q}^T & (\frac{k-1}{k})^2\mathbf{A}_{\mathbf{m},k-1} 
                        \end{bmatrix} \\
                        &+ \begin{bmatrix}
                        \mathbf{J}_{\mathbf{n},\mathbf{p}_k,k} \mathbf{A}_{\mathbf{p}_k} \mathbf{J}_{\mathbf{n},\mathbf{p}_k,k}^T & \frac{1}{k}\mathbf{J}_{\mathbf{n},\mathbf{p}_k,k} \mathbf{A}_{\mathbf{p}_k} \\
                         \frac{1}{k}\mathbf{A}_{\mathbf{p}_k} \mathbf{J}_{\mathbf{n},\mathbf{p}_k,k}^T & \frac{1}{k^2}\mathbf{A}_{\mathbf{p}_k}
                        \end{bmatrix}
        \end{aligned}
        \label{eq_cnqkk1}
\end{equation}



In contrast to methods like VoxelMap \cite{yuan2022efficient}, \eqref{eq_cnqkk1} offers a significant computational advantage by achieving a constant time complexity of  $\mathcal{O} (1)$ . 
This translates to faster execution, making it well-suited for real-time applications.

\section{Integration with LIO}
\label{sec:loglio2}
In this section, we integrate the proposed point uncertainty model along with LUFA into our prior work \cite{huang2023log}, namely LOG-LIO2. 
The system state $\mathbf{x}$ is as follows:
\begin{equation}
        \mathbf{x} = \left[^\mathcal{W}\mathbf{R}^T_\mathcal{I}\ ^\mathcal{W}\mathbf{p}^T_\mathcal{I}\ ^\mathcal{W}\mathbf{v}^T_\mathcal{I}\ \mathbf{b}^T_\omega\ \mathbf{b}^T_a\ ^\mathcal{W}\mathbf{g}^T\right]
        \label{eq_x}
\end{equation}
where $^\mathcal{W}\mathbf{R}^T_\mathcal{I}$, $^\mathcal{W}\mathbf{p}^T_\mathcal{I}$, and $^\mathcal{W}\mathbf{v}^T_\mathcal{I}$ are the orientation, position, and velocity of IMU in the world frame.
$\mathbf{b}^T_\omega$ and $\mathbf{b}^T_a$ are gyroscope and accelerometer bias respectively.
$^\mathcal{W}\mathbf{g}^T$ is the known gravity vector in the world frame.

\begin{figure}[!ht]
        \centering
        {
                \begin{minipage}[b]{0.46\textwidth}
                        \includegraphics[width=1\textwidth]{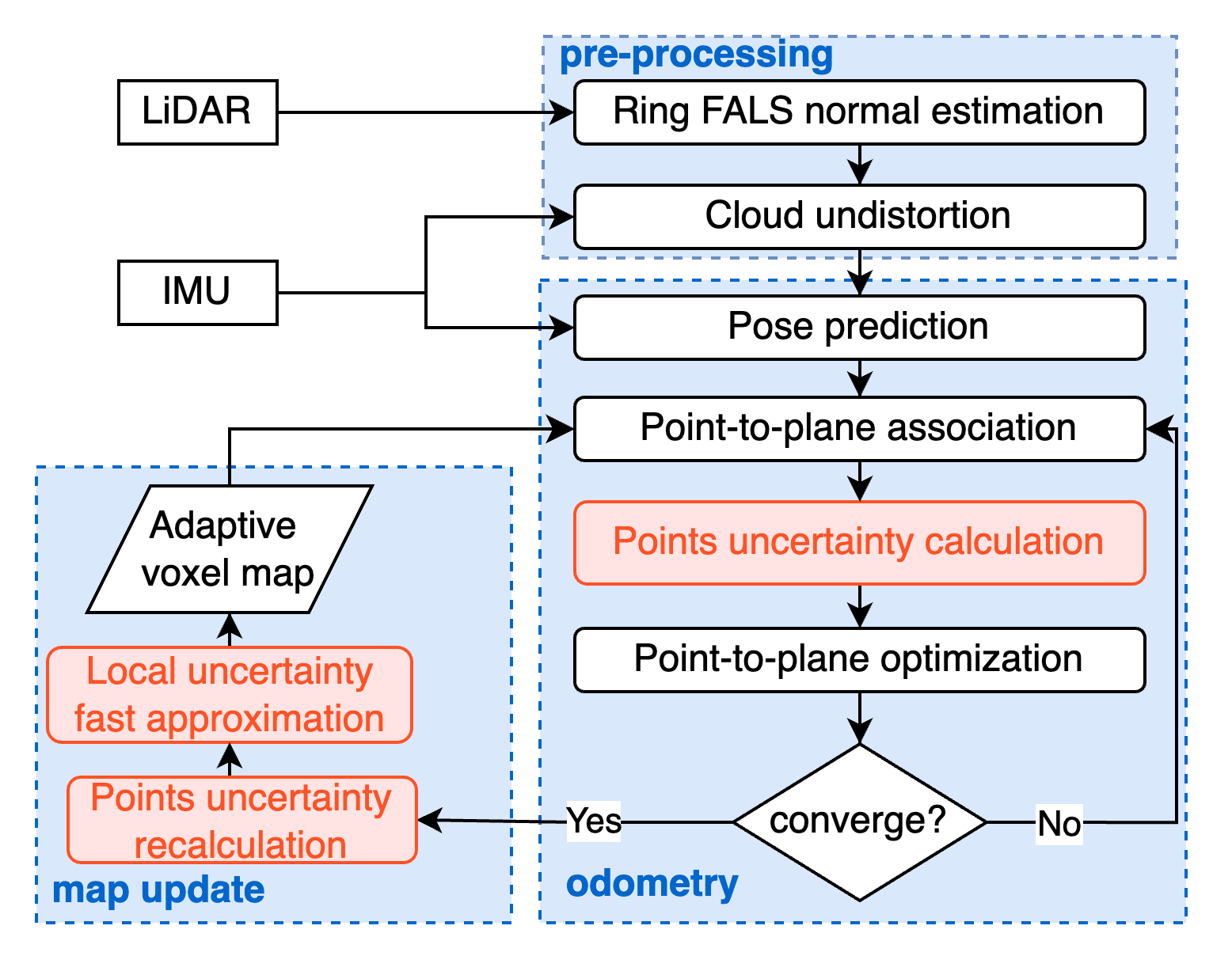}
                \end{minipage}
        }
        \caption{System overview of LOG-LIO2 with enhancements marked in red.}
        \label{fig_lio2}
\end{figure}

The system structure is depicted in \Reffig{fig_lio2}. 
It comprises three modules: pre-processing, odometry, and map update.
We employ adaptive voxelization from VoxelMap \cite{yuan2022efficient} for map management, leveraging its efficient data association and incorporating the computation of within-voxel covariance.

\subsection{Pre-processing}
For a point $\mathbf{p}_i$ within the new input scan, we first estimate its normal $\mathbf{n}_{r_i}$ using Ring FALS \cite{huang2023log}.
Subsequently, point cloud undistortion is performed to compensate for LiDAR motion based on IMU measurements \cite{xu2022fast}.

\subsection{Odometry}
By incorporating the IMU measurements from the previous scan as a prediction $\widehat{\mathbf{x}}_{t}$ for the current scan, we transform each point into the world frame. 
Each point is then assigned to the voxel it resides in if the points within that voxel satisfy planar geometry (indicated by the minimum eigenvalue below a threshold).
Suppose $\mathbf{p}_i$ associates with the $l$-th voxel $v_l$, we compute the point-to-plane distance as:
\begin{equation}
        \mathbf{z}_i = \mathbf{n}^T_{v_l}(\mathbf{p}_i - \mathbf{m}_{v_l}).
\end{equation}
where $\mathbf{m}_{v_l}$ and $\mathbf{n}_{v_l}$ denote the center and normal of $v_l$, respectively.
Notably, $\mathbf{n}_{v_l}$ differs from $\mathbf{n}_{r_i}$ in that it incorporates geometric information from all map points within $v_l$, making it more suitable for the incident angle calculation \eqref{eq_incident},
whereas $\mathbf{n}_{r_i}$ focuses on the local geometry of individual points in the scan. 
Both $\mathbf{n}_{v_l}$ and $\mathbf{n}_{r_i}$ contribute to roughness estimation \eqref{eq_roughness}.

We denote the propagated state and covariance by $\widehat{\mathbf{x}}_{t}$ and $\widehat{\mathbf{P}}_{t}$ respectively.
By incorporating the prior distribution and stacking all the point-to-plane associations,
we obtain the maximum a-posterior estimate (MAP) as follows \cite{xu2022fast}:
\begin{equation}
        \begin{aligned}
                \underset{\widetilde{\mathbf{x}}_t^\kappa}{min} ( \|\mathbf{x}_{t}\boxminus \widehat{\mathbf{x}}_t\Vert^2_{\widehat{\mathbf{P}}_t^{-1}} +\sum_{i\in plane} \|\mathbf{z}_i^\kappa + \mathbf{H}_i^\kappa \widetilde{\mathbf{x}}_t^\kappa \Vert _{\mathbf{A}_{\mathbf{p}_i,\mathbf{n}_{v_l},\mathbf{m}_{v_l}}^{-1}}^2)
        \end{aligned}
        \label{eq_map}
\end{equation}
where $\boxminus$ computes the difference  between $\mathbf{x}_t$ and $\widehat{\mathbf{x}}_t$ in the local tangent space of $\mathbf{x}_t$,
$\widetilde{\mathbf{x}}_t^\kappa$ is the error of the $\kappa$-th iterate update at time $t$, $\mathbf{H}_i^\kappa$ is the Jacobian matrix w.r.t. $\widetilde{\mathbf{x}}_t^\kappa$.
The point-to-plane distance is weighted by ${\mathbf{A}_{\mathbf{p}_i,\mathbf{n}_{v_l},\mathbf{m}_{v_l}}^{-1}}$, 
which incorporates point, normal, and center uncertainties.
We initialize the point-wise uncertainty in the LiDAR frame as detailed in \Refsec{sec:uncer_model} and then transform it to the world frame.
Normal and center uncertainties are estimated by LUFA using map points in the corresponding voxel. 
We employ iEKF \cite{xu2022fast} to optimize the system state.

\subsection{Map update}
After optimization, each point is assigned to the corresponding voxel with the updated state.
A crucial aspect of this module lies in propagating the uncertainty from the scan points to the voxel center, as well as its eigenvalues and eigenvectors. 
The uncertainty of points is first recalculated based on the updated state.
Apart from increment magnitude checks (see \Refsec{sec:lufa}), 
we employ LUFA considering two additional factors:
\subsubsection{Map stability}
To strike a balance between initial map instability and computational efficiency, 
we employ LUFA after points within a voxel exceed a predefined threshold $n_{min}$.

\subsubsection{Error accumulation}
The fast approximation introduces small errors with each iteration.
To alleviate the accumulation of these errors, we limit the continuous application of LUFA to a maximum of $n_{ct}=100$ iterations.

Beyond the above conditions, the rigorous update forms \eqref{eq_dlk_balm} and \eqref{eq_cvv} are employed.
Additionally, once the point count reaches $n_{max}$, we fix the voxel update using rigorous forms to ensure the accuracy of the local geometric information. 
This adaptive strategy effectively balances stability and efficiency during map updates.

\section{EXPERIMENTAL RESULTS}
\label{sec:experimental_results}
\subsection{Experimental Settings}
The experiment focuses on the following two research questions:
\begin{itemize}
        \item
        Can LUFA approximate the uncertainty computed in rigorous form?
        \item
        Can LOG-LIO2 improve accuracy and efficiency by incorporating our point uncertainty model and LUFA?
\end{itemize}
We first validate LUFA in a simulation environment and then further test the performance of LOG-LIO2 on real-world datasets.
Our workstation runs with Ubuntu 18.04, equipped with an Intel Core Xeon(R) Gold 6248R 3.00GHz processor and 32GB RAM.

\subsection{LUFA Experiments}
\begin{figure}[!ht]
        \centering
        {
                \begin{minipage}[b]{0.25\textwidth}
                        \includegraphics[width=1\textwidth]{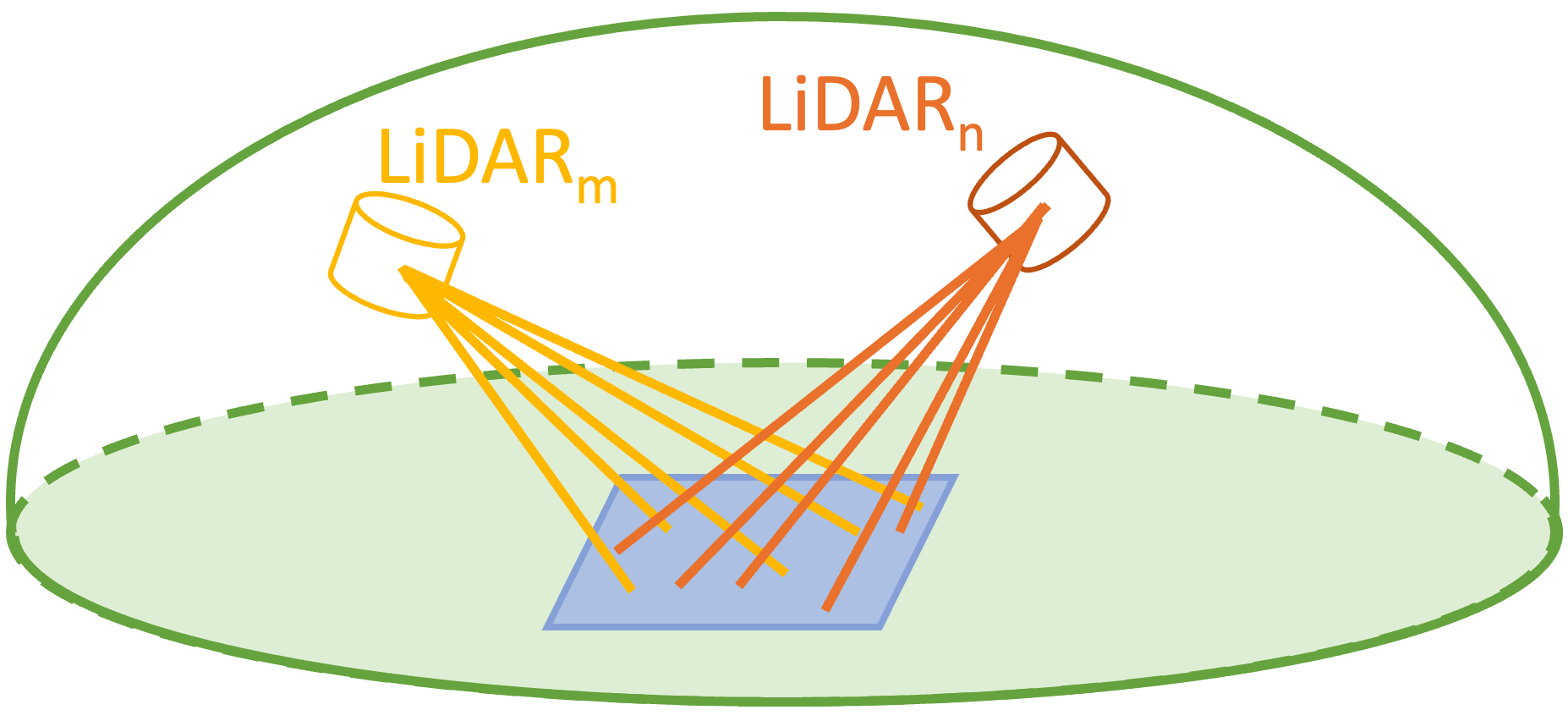}
                \end{minipage}
        }
        \caption{Illustration of the simulation environment.}
        \label{fig_simu}
\end{figure}
To assess the accuracy and efficiency of LUFA, we benchmark it against BALM \cite{liu2021balm}.
\Reffig{fig_simu} illustrates the validation environment, simulating a $20m \times 20m$ plane with 20 LiDARs, each LiDAR captures 50 points on the plane.
All LiDARs are positioned on the same side of the plane and remain within a 100m radius of its center. 
The center and normal of the plane, the poses of the LiDARs, and the coordinates of the sampled points on the plane are all randomly generated.
To mimic real-world scenarios, we corrupt all these parameters with Gaussian noise, accounting for factors such as the plane's unevenness, inaccuracies in the LiDAR poses, and measurement errors.
The point uncertainty is calculated as detailed in \Refsec{sec:uncer_model}. 
We set $n_{min}=200$ to define when LUFA is triggered.

\subsubsection{Accuracy}
\begin{figure}[!ht]
        \centering
        {
                \begin{minipage}[b]{0.46\textwidth}
                        \includegraphics[width=1\textwidth]{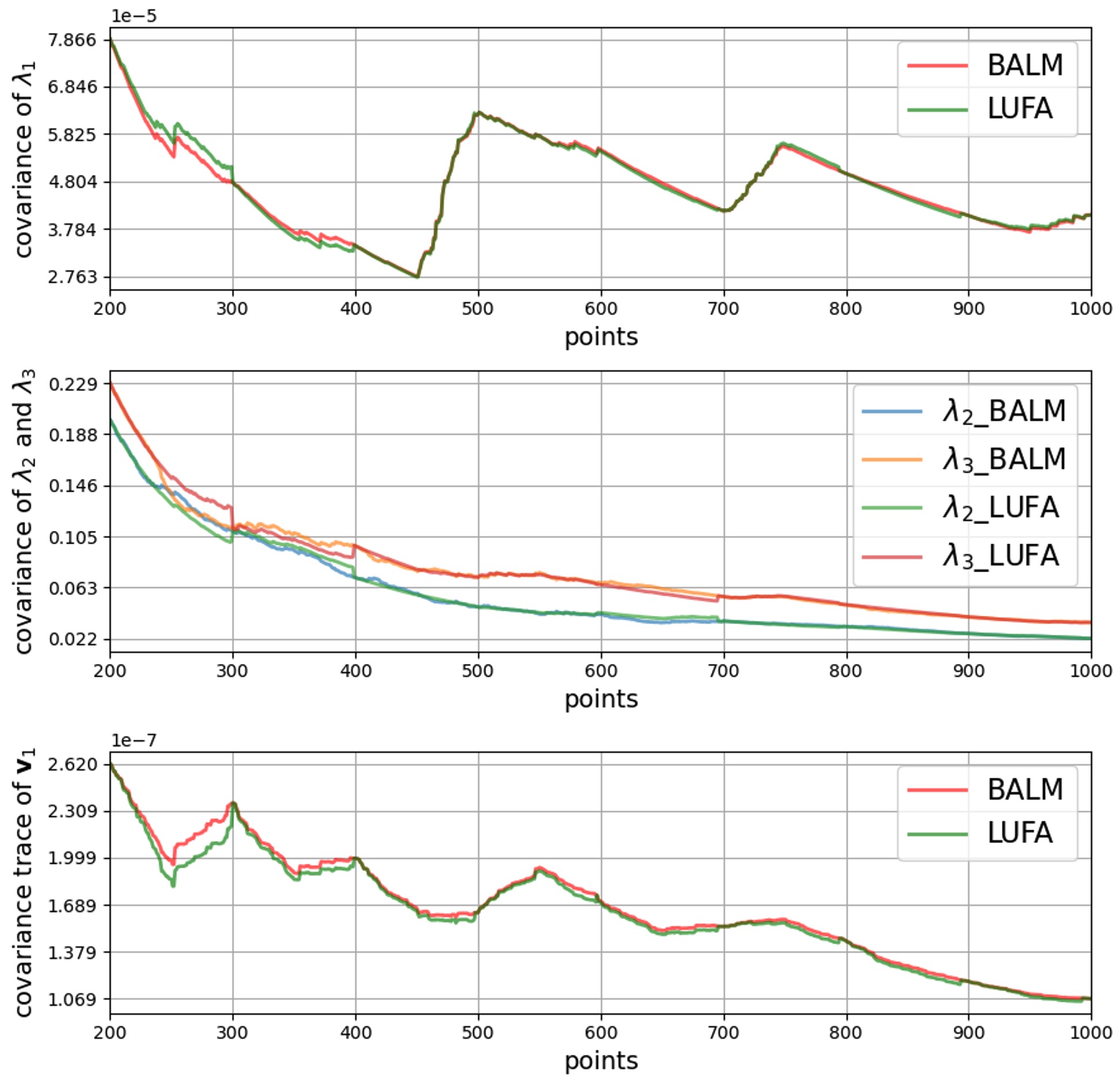}
                \end{minipage}
        }
        \caption{Comparison of covariance calculated by LUFA and BALM.}
        \label{fig_lambda_n}
\end{figure}

\Reffig{fig_lambda_n} illustrates the covariance of the $\lambda_j$ and the covariance trace of $\mathbf{v}_1$ computed by LUFA and BALM. 
As the covariance of the eigenvectors involves matrices, we compare the trace of the covariance matrices in our experiments, which provides a measure of the overall magnitude of the covariance.
The results indicate that the covariance computed by LUFA is close to that of BALM.
This closeness in covariance values demonstrates the effectiveness of LUFA in approximating the uncertainty propagation while maintaining a reasonable level of accuracy.
\begin{figure}[!ht]
        \centering
        {
                \begin{minipage}[b]{0.47\textwidth}
                        \includegraphics[width=1\textwidth]{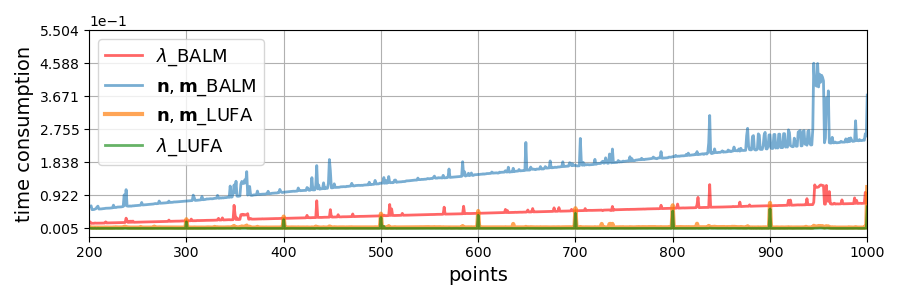}
                \end{minipage}
        }
        \caption{Time consumption for computing the covariance.
        $\mathbf{n}$, $\mathbf{m}$ represent the covariance of the normal and the center.}
        \label{fig_time_lambda}
\end{figure}
\subsubsection{Efficiency}
We further plot the processing time w.r.t the number of points for better evaluation in \Reffig{fig_time_lambda}.
As expected, BALM exhibits linear time complexity $\mathcal{O} (n)$, resulting in processing time increases proportionally with the number of points. 
In contrast, the processing time of LUFA remains nearly constant, adhering to  $\mathcal{O} (1)$ time complexity.
The observed peaks in LUFA's processing time occur at intervals of 100 points, which stems from our imposed limitation on the maximum number of consecutive LUFA executions.
Utilizing the incremental center \eqref{eq_mk} and covariance \eqref{eq_sk} ensures that even at peak processing times, LUFA remains computationally more efficient than BALM.

By comparing the results obtained using LUFA with those from BALM, we demonstrate that LUFA achieves comparable accuracy while significantly improving computational efficiency.

\subsection{LIO Experiments}
To evaluate the efficacy of LOG-LIO2 in real-world scenarios, we employ the M2DGR dataset \cite{yin2021m2dgr}.
This dataset collects data on a ground platform equipped with Velodyne-32 LiDAR and presents significant challenges for our previous method, LOG-LIO. 
To isolate the impact of specific components within LOG-LIO2, we introduce LOG2-i.
LOG2-i differs from LOG-LIO2 in its map uncertainty update mechanism, employing the formulation outlined in BALM instead of the LUFA approach. 
We compare the performance of LOG-LIO2 against two leading LIO methods: FAST-LIO2 \cite{xu2022fast} and PV-LIO
\footnote{https://github.com/HViktorTsoi/PV-LIO}.
PV-LIO is the reimplementation of VoxelMap \cite{yuan2022efficient} integrating IMU, showing promising results in practice.
Both PV-LIO and VoxelMap use BALM's formulation for uncertainty propagation.

To ensure a fair comparison, LOG-LIO2, LOG2-i, and PV-LIO use identical parameters across all dataset sequences: maximum voxel size of 1.6m and maximum octree layer of 3, resulting in a minimum voxel size of 0.2m. 
FAST-LIO2 and LOG-LIO are run with a map resolution of 0.4m using ikd-tree. 
Additionally, we set $n_{min}$ and $n_{max}$ to 200 and 1000 respectively, defining the thresholds for when LUFA is triggered and when it is no longer necessary.
Due to the instability in the RTK signal, the first and last 100 seconds of sequences \emph{street\_07} and \emph{street\_10} are excluded from the evaluation. 
Note that all the LIO systems above perform scan-to-map registration by minimizing point-to-plane distances and loop closure was disabled for all experiments.

\subsubsection{Accuracy Evaluation}

\Reftab{tab_m2dgr} reports the root-mean-square-error (RMSE) of absolute trajectory error (ATE).
LOG-LIO2 and LOG2-i stand out as the most accurate LIO systems, exhibiting comparable performance, with LOG-LIO and PV-LIO following behind. 
The advantages of  LOG-LIO2, LOG2-i, and PV-LIO stem from a combination of an adaptive voxel map and uncertainty-weighted point-to-plane distance for scan-to-map registration. 
This approach offers a more accurate geometric representation, leading to minimal registration errors.
Conversely, LOG-LIO and FAST-LIO2 depend on a fixed-scale tree structure for map management and implement isotropic noise-weighted point-to-plane distance. 
Notably, LOG-LIO distinguishes itself from FAST-LIO2 by incrementally maintaining normal and point distribution within map nodes, thereby enabling the capture of geometric complexities.

\begin{figure}[!ht]
        \centering
        {
                \begin{minipage}[b]{0.48\textwidth}
                        \includegraphics[width=1\textwidth]{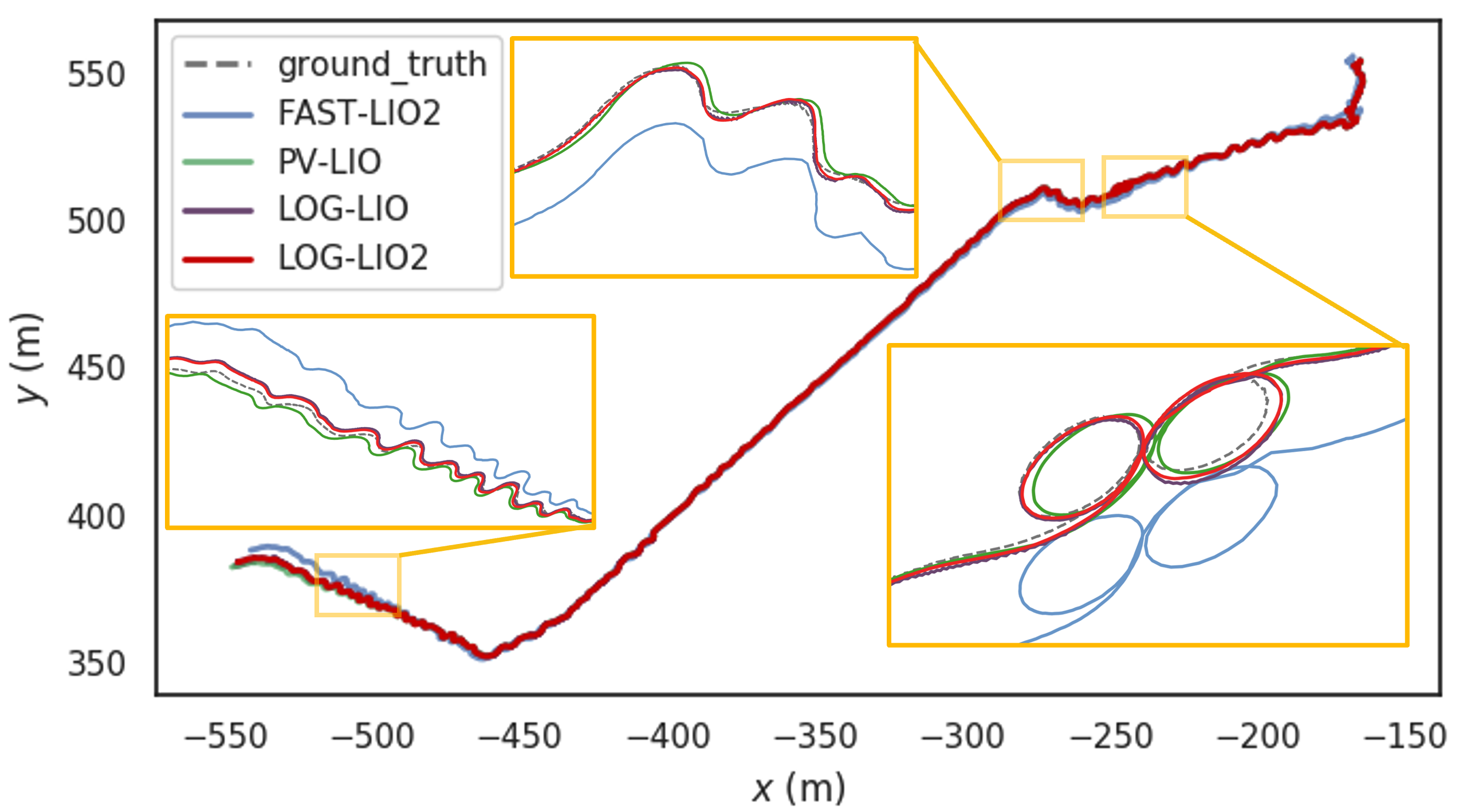}
                \end{minipage}
        }
        \caption{Localization estimates in sequence \emph{street\_07} of the M2DGR dataset.
                 }
        \label{fig_traj}
\end{figure}

\begin{table}[!ht]
        \caption{The Translation RMSE(m) Results of Pose Estimation Comparison on the M2DGR Dataset}
        \begin{tabular}{c|c|c|c|c|c}
                \toprule
                           & \scriptsize{LOG-LIO2} & \scriptsize{LOG2-i}& \scriptsize{LOG-LIO} & \scriptsize{PV-LIO}  & \scriptsize{FAST-LIO2} \\
                \midrule
                walk\_01   & 0.131          & 0.132          & \underline{0.117}    & 0.135          & \textbf{0.112} \\
                door\_01   & \underline{0.264}  & \textbf{0.262}    & 0.266          & \textbf{0.262} & 0.271          \\
                door\_02   & 0.197          & 0.197          & \underline{0.196}    & \textbf{0.194} & 0.200          \\
                street\_01 & 0.343          & 0.303          & \underline{0.291}    & 0.439          & \textbf{0.272} \\
                street\_02 & \underline{2.625}    & \textbf{2.541} & 3.252          & 3.540          & 2.754          \\
                street\_03 & 0.104          & 0.104          & \textbf{0.092} & \underline{0.093}    & 0.106          \\
                street\_04 & 1.032          & 1.079          & \underline{0.697}    & 1.081          & \textbf{0.552} \\
                street\_05 & 0.426          & \underline{0.370}    & \textbf{0.306} & 0.543          & 0.377          \\
                street\_06 &  \underline{0.355} & \textbf{0.338} & \underline{0.355}    & 0.494          & 0.434          \\
                street\_07 & \underline{0.358}    & \textbf{0.339} & 0.422          & 0.651          & 3.512          \\
                street\_08 & 0.166          & 0.156          & \underline{0.140}    & \textbf{0.138} & 0.170          \\
                street\_09 & \underline{1.822}    & 1.861          & 2.380          & \textbf{1.678} & 3.648          \\
                street\_10 & \textbf{0.314} & 0.369          & \underline{0.349}    & 0.464          & 0.956          \\                \midrule
                mean       & \underline{0.626}     & \textbf{0.619}      & 0.682               & 0.747                & 1.028                  \\
                \bottomrule
        \end{tabular}
        \label{tab_m2dgr}
        \begin{tablenotes}
                \footnotesize
                \item The best and second-best results are bolded and underlined respectively.
        \end{tablenotes}
\end{table}
LOG2-i and LOG-LIO2 outperform PV-LIO in most sequences due to our comprehensive point uncertainty model. 
PV-LIO's model suffers from limitations: constant uncertainty in the ray direction
and neglecting surface geometry.
In contrast, our model considers incident angle and surface roughness, providing a more precise environmental representation. 
This advantage is evident in large outdoor environments like the \emph{street} sequences. 
In smaller indoor spaces like \emph{door} sequences, all systems achieve comparable performance due to lower point uncertainty and well-defined structural planes. 
\Reffig{fig_traj} compares the trajectories in the \emph{street\_07} sequence, highlighting the performance difference.

\subsubsection{Efficiency Evaluation}
\Reftab{tab_street_time} details the processing times of LOG-LIO2, LOG2-i, and PV-LIO for the \emph{street} sequence, 
highlighting map update and runtime behavior across different thread configurations. 
Note that Ring FALS introduces an additional 6ms overhead within the pre-processing module of LOG-LIO2 and LOG2-i.
Despite this overhead, comparing the average time consumption with 4 threads, LOG2-i exhibits a slightly shorter duration than PV-LIO, 
mainly benefiting from the fast uncertainty calculation (\ref{sec:fast_uncer}), increment center, and covariance (\ref{sec:iCov}).
LOG-LIO2 stands out for its efficiency with LUFA, 
significantly reducing map update times compared to PV-LIO, operating nearly half the time of PV-LIO with a single thread and two-thirds the time with four threads.
\begin{table}[!ht]
        \caption{The Average Time Consumption(ms) of street Sequence in The Experiments}
        \begin{tabular}{c|cc|cc|cc}
                \toprule
                   & \multicolumn{2}{c|}{LOG-LIO2} & \multicolumn{2}{c|}{LOG2-i}   & \multicolumn{2}{c}{PV-LIO} \\
                   thread(s) & 1       & 4   &1  &4 & 1      & 4   \\
                   \midrule
                   map update & 19.13 & \textbf{9.29} & 28.67 & \underline{11.92} & 37.98        & 14.56       \\
                   total      & 59.20  & \textbf{50.77} & 68.68 & \underline{52.83} & 80.27        & 59.47       \\
                   \bottomrule
        \end{tabular}
        \label{tab_street_time}
        \begin{tablenotes}
                \footnotesize
                \item The best and second-best results are bolded and underlined respectively.
        \end{tablenotes}
\end{table}

\section{CONCLUSION}
\label{sec:conclusion}
This paper presents a comprehensive point uncertainty model alongside a fast calculation method utilizing the projection operator. 
This model incorporates not only the range and bearing uncertainty of LiDAR but also the uncertainty arising from incident angle and surface roughness.
Furthermore, we derive the incremental Jacobian matrices of the eigenvalues and eigenvectors, enabling the application of the local uncertainty fast approximation (LUFA).
The accuracy and efficiency of our formulations are first demonstrated through simulation experiments, benchmarking against the rigorous form derived by BALM. 
Subsequently, we integrate all aforementioned methods into the LIO system LOG-LIO2.
Ablation experiments conducted on a public dataset validate the accuracy and efficiency of LOG-LIO2 compared to state-of-the-art LIO systems.

\appendices
\section{}
\subsection{Consistency of Projection Operator with $\mathbb{S}^2$ Perturbation}
\label{sec:p_po}
In \cite{yuan2021pixel, liu2021balm, yuan2022efficient}, the point uncertainty model considering range and bearing is derived from the perturbation of the true range $d_i^{\text{gt}}$  and true bearing $\boldsymbol{\omega}_i^{\text{gt}}$ as:
\begin{equation}
        \begin{split}
                \mathbf p_i^{\text{gt}} &= d_i^{\text{gt}} \boldsymbol{\omega}_i^{\text{gt}} = \left( d_i + \sigma_{d_i} \right) \left( \boldsymbol{\omega}_i \boxplus_{\mathbb{S}^2} \boldsymbol{\sigma}_{\boldsymbol{\omega}_i} \right)        
        \end{split}
\end{equation}
where $d_i$ and $\boldsymbol{\omega}_i$ is range and bearing measurements respectively,
$\boxplus$-operation encapsulated in $\mathbb{S}^2$ \cite{he2021kalman},
$\sigma_{d_i} \sim \mathcal{N}(0, \Sigma_{d_i})$ and $\boldsymbol{\sigma}_{\boldsymbol{\omega}_i} \sim \mathcal{N}(\mathbf 0_{2\times 1}, \boldsymbol{\Sigma}_{\boldsymbol{\omega}_i})$ are range and bearing noise, respectively.
The Jacobian matrix of $\mathbf p_i$ w.r.t. $d_i$ and $\boldsymbol{\sigma}_{\boldsymbol{\omega}_i}$ can be further obtained as
 $                \mathbf{J}_{d_i,\boldsymbol{\omega}_i}= 
 \begin{bmatrix}
        \boldsymbol{\omega}_i  & -d_i\left\lfloor \boldsymbol{\omega}_i \right\rfloor_\times\mathbf{N}(\boldsymbol{\omega}_i)
        \end{bmatrix}$,
where $\lfloor  \   \rfloor_\times$ denotes the skew-symmetric matrix mapping the cross product,
and $\mathbf N(\boldsymbol{\omega}_i) = [\mathbf N_1\ \mathbf N_2] \in \mathbb{R}^{3 \times 2}$ is the orthogonal basis of the tangent plane at $\boldsymbol{\omega}_i$.
Then the covariance of the point leads to:
\begin{equation}
        \begin{aligned}
                \mathbf{A}_{\omega_i} &= 
                \mathbf{J}_{d_i,\boldsymbol{\omega}_i}
                \begin{bmatrix}
                \sigma_{d_i}^2 & \mathbf{0}_{\text{1x2}}
                \\
                \mathbf{0}_{\text{2x1}} & \boldsymbol{\Sigma}_{\boldsymbol{\omega}_i}^2 
                \end{bmatrix} 
                \mathbf{J}_{d_i,\boldsymbol{\omega}_i}^T 
        \end{aligned}
\end{equation}
Despite constructing $\mathbf N(\boldsymbol{\omega}_i)$ arbitrarily in the tangent plane of $\boldsymbol{\omega}_i$ is feasible, this approach incurs a computational cost. 
We show that utilization of the projection operator simplifies the computation while maintaining consistency:
\begin{equation}
        \begin{aligned}
                \mathbf{A}_{\omega_i} &= \sigma_{d_i}^2 \boldsymbol{\omega}_i \boldsymbol{\omega}_i^T + d_i^2 \sigma_\omega^2 \left\lfloor \boldsymbol{\omega}_i \right\rfloor_\times \mathbf{N}(\boldsymbol{\omega}_i) \mathbf{I}_{\text{2x2}} \mathbf{N}(\boldsymbol{\omega}_i)^T \left\lfloor \boldsymbol{\omega}_i \right\rfloor_\times^T \\
                &= \sigma_{d_i}^2 \boldsymbol{\omega}_i \boldsymbol{\omega}_i^T + d_i^2 \sigma_\omega^2 \left\lfloor \boldsymbol{\omega}_i \right\rfloor_\times (-\left\lfloor \boldsymbol{\omega}_i \right\rfloor_\times) \\
                &= \sigma_{d_i}^2 \boldsymbol{\omega}_i \boldsymbol{\omega}_i^T + d_i^2 \sigma_\omega^2 (\mathbf{I}_{3\times3} - \boldsymbol{\omega}_i \boldsymbol{\omega}_i^T).   
        \end{aligned}
        \label{eq_cov_voxelmap}
\end{equation}
Therefore, by eliminating the need for constructing  $\mathbf N(\boldsymbol{\omega}_i)$, our approach results in a more efficient method.

\subsection{Proof of The Incremental Jacobian of Eigenvalue}
\label{sec:p_ije}

Similar to BALM \cite{liu2021balm}, in this section, eigenvectors are viewed as constant.
Multiplying \eqref{eq_AkAk1} by $\mathbf{v}_{j,k-1}^T$ on the left and $\mathbf{v}_{j,k}$ on the right, and combine with (\ref{eq_pca}), we get:
\begin{equation}
        \begin{aligned}
                \lambda_{j,k} \mathbf{v}_{j,k-1}^T \mathbf{v}_{j,k} &= \frac{k-1}{k} (\lambda_{j,k-1} \mathbf{v}_{j,k-1}^T \mathbf{v}_{j,k} + 
                        \frac{1}{k} \mathbf{v}_{j,k-1}^T\mathbf{D} \mathbf{v}_{j,k}) \\                        
                \lambda_{j,k} &= \frac{k-1}{k}\lambda_{j,k-1} + \frac{k-1}{k^2 \cos\theta_j}\mathbf{v}_{j,k-1}^T \mathbf{D} \mathbf{v}_{j,k},
        \end{aligned}
        \label{eq_lklk1}
\end{equation}
where $\theta_j$ is the angle between $\mathbf{v}_{j,k}$ and $\mathbf{v}_{j,k-1}$.

Then the partial derivative of the updated eigenvalue $\lambda_{j,k}$ w.r.t. $\mathbf{p}_i$ is given by:
\begin{equation}
        \begin{aligned}
                \frac{\partial \lambda_{j,k}}{\partial \mathbf{p}_i} = \frac{k-1}{k}\frac{\partial \lambda_{j,k-1}}{\partial \mathbf{p}_i}
                + \frac{k-1}{k^2 \cos\theta_j} \frac{\partial \mathbf{v}_{j,k-1}^T\mathbf{D} \mathbf{v}_{j,k}}{\partial \mathbf{p}_i}.
        \end{aligned}
        \label{eq_dlk_dp}
\end{equation}
Given the value of ${\partial \lambda_{j,k-1}}/{\partial \mathbf{p}_i}$ in the last update, only the second term of the above equation needs to be solved.

For $i = 1,...,k-1$, the second term of (\ref{eq_dlk_dp}) is:
\begin{equation}
        \begin{aligned}
                &\frac{k-1}{k^2 \cos\theta_j} \frac{\partial \mathbf{v}_{j,k-1}^T\mathbf{D} \mathbf{v}_{j,k}}{\partial \mathbf{p}_i} \\
                                =& \frac{k-1}{k^2 \cos\theta_j}[-\frac{\mathbf{v}_u^T\mathbf{v}_{j,k-1} \mathbf{v}_{j,k}^T}{(k-1)} -\frac{\mathbf{v}_u^T\mathbf{v}_{j,k}\mathbf{v}_{j,k-1}^T}{(k-1)}]\\
                                =& -\frac{d_u}{k^2 \cos\theta_j}(\cos\varphi_{j,k-1}\mathbf{v}_{j,k}^T + \cos\varphi_{j,k}\mathbf{v}_{j,k-1}^T)    
                \end{aligned}
        \label{eq_a_dlk_dpj}
\end{equation}
where $d_u$ is the distance between $\mathbf{p}_k$ and $\mathbf{m}_{k-1}$.
$\varphi_{j,k}$ and $\varphi_{j,k-1}$ are the angles from the direction of $\mathbf{v}_u$ to the $j$-th eigenvector $\mathbf{v}_{j,k}$ derived from $k$ points and the $j$-th eigenvector $\mathbf{v}_{j,k-1}$ derived from $k-1$ points, respectively.

For $\mathbf{p}_k$, which is independent of $\lambda_{j,k-1}$, (\ref{eq_dlk_dp}) simplifies to:
\begin{equation}
        \begin{aligned}
                \frac{\partial \lambda_{j,k}}{\partial \mathbf{p}_k} &= \frac{k-1}{k^2\cos\theta_j}
                        \frac{\partial \mathbf{v}_{j,k-1}^T \mathbf{D} \mathbf{v}_{j,k}}{\partial \mathbf{p}_k}    \\
                &= \frac{k-1}{k^2\cos\theta_j} 
                        (\mathbf{v}_u^T\mathbf{v}_{j,k-1} \mathbf{v}_{j,k}^T +  \mathbf{v}_u^T \mathbf{v}_{j,k}\mathbf{v}_{j,k-1}^T) \\
                &= \frac{d_u(k-1)}{k^2 \cos\theta_j}(\cos\varphi_{j,k-1}\mathbf{v}_{j,k}^T + \cos\varphi_{j,k}\mathbf{v}_{j,k-1}^T)      .
          \end{aligned}
        \label{eq_a_dlk}
\end{equation}

\subsection{Proof of The Incremental Jacobian of Eigenvectors}
\label{sec:p_ijevectors}
We first revisit the derivations of BALM \cite{liu2021balm} from \eqref{eq_vkvk1} to \eqref{eq_ckq}.
Reformulating (\ref{eq_pca}), we get:
\begin{equation}
        \begin{aligned}
                \mathbf{\Lambda}  = \mathbf{V}^T \mathbf{A}  \mathbf{V};
                \mathbf{V} \mathbf{\Lambda} = \mathbf{A}  \mathbf{V};
                 \mathbf{\Lambda} \mathbf{V}^T =  \mathbf{V}^T \mathbf{A}.
        \end{aligned}
        \label{eq_vkvk1}
\end{equation}

Utilizing the orthogonality property $\mathbf{V}^T\mathbf{V}=\mathbf{I}$, we can further obtain:
\begin{equation}
        \begin{aligned}
                \mathbf{V}^{T} \frac{\partial\mathbf{V}}{\partial q_i}+ \left( \frac{\partial\mathbf{V}}{\partial q_i}  \right)^T \mathbf V = \mathbf{0}.
        \end{aligned}
        \label{eq_vtv}
\end{equation}

Denoting an element of $\mathbf{p}_i$ by $q_i$ and combining (\ref{eq_vkvk1}), the derivative of $\mathbf{\Lambda}_k$ w.r.t. $q_i$ leads to:
\begin{equation}
        \begin{aligned}
                &\frac{\partial\mathbf{\Lambda}_k}{\partial q_i} 
                = \mathbf{V}_k^{T}\frac{\partial\mathbf{A}_k}{\partial q_i}\mathbf{V}_k 
                + (\frac{\partial\mathbf{V}_k}{\partial q_i})^{T}{\mathbf{A}_k}\mathbf{V}_k
                +\mathbf{V}_k^{T}{\mathbf{A}_k}\frac{\partial\mathbf{V}_k}{\partial q_i}
                \\
                &=\mathbf{V}_k^{T}\frac{\partial\mathbf{A}_k}{\partial q_i}\mathbf{V}_k 
                + {\mathbf{\Lambda}_k} \underbrace{\mathbf{V}_k^{T} \frac{\partial\mathbf{V}_k}{\partial q_i}}_{(\mathbf C^{q})_k} 
                + \underbrace{\left( \frac{\partial\mathbf{V}_k}{\partial q_i}  \right)^T  \mathbf V_k}_{(\mathbf{C}^{q})^T_k} \mathbf \Lambda_k.
        \end{aligned}
        \label{eq_dvdq}
\end{equation}

Given that $\mathbf{\Lambda}$ is diagonal, the off-diagonal elements in \eqref{eq_dvdq} can be combined with \eqref{eq_vtv} to yield:
\begin{align*}
        0 = \mathbf{v}_{m,k}^T  \frac{\partial\mathbf{A}_k}{\partial q}\mathbf{v}_{n,k} + \lambda_{m,k} (\mathbf{C}^{q}_{m,n})_k - (\mathbf{C}^{q}_{m,n})_k\lambda_{n,k}.
\end{align*}
Here, $(\mathbf{C}^{q}_{m,n})_k$ represents the element at the $m$-th row and $n$-th column of $(\mathbf{C}^{q})_k$, defined as:
\begin{align} \label{eq_ckq}
        (\mathbf{C}^{q}_{m,n})_k &=\left\{\begin{aligned}
         \frac{1}{(\lambda_{n} - \lambda_{m})_k} \mathbf{v}_{m,k}^T \frac{\partial \mathbf{A}_k }{\partial  q} \mathbf{v}_{n,k} , m\neq n \\
        0  \qquad\qquad, m=n
        \end{aligned}\right .
\end{align}
To derive the incremental form of the matrix $\mathbf{C}$ as described above, we derive the partial derivatives in (\ref{eq_ckq}) using (\ref{eq_AkAk1}):
\begin{equation}
        \begin{aligned}
                \frac{\partial \mathbf{A}_k}{\partial q} = \frac{k-1}{k} \frac{\partial \mathbf{A}_{k-1}}{\partial q} + \frac{k-1}{k^2} \frac{\partial \mathbf{D}}{\partial q}.
        \end{aligned}
        \label{eq_dbdq}
\end{equation}

The specific form of $(\mathbf{C}^{q}_{m,n})_{k-1}$, $m \neq n $, derived from BALM is given by:
\begin{equation}
        \begin{aligned}
                (\mathbf{C}^{q}_{m,n})_{k-1} = \frac{1}{(\lambda_{n} - \lambda_{m})_{k-1}} \mathbf{v}_{m,k-1}^T \frac{\partial \mathbf{A}_{k-1} }{\partial  q} \mathbf{v}_{n,k-1}.
        \end{aligned}
        \label{eq_dak1}
\end{equation}

Similar to (\ref{eq_a_dlk_dpj}), for $j=1,...,k-1$, we can further obtain:
\begin{equation}
        \begin{aligned}
                \mathbf{v}_{m,k}^T \frac{\partial \mathbf{D} }{\partial \mathbf{p}_j} \mathbf{v}_{n,k}
        = -\frac{d_u}{k-1} (\cos\varphi_{m}\mathbf{v}_{n}^T + \cos\varphi_{n}\mathbf{v}_{m}^T)_k
\end{aligned}
        \label{eq_vdpjv}
\end{equation}
where $\varphi_{m}$ and $\varphi_{n}$ denote angles from the direction of $\mathbf{v}_u$ to $\mathbf{v}_{m}$ and $\mathbf{v}_{n}$, respectively.

Substituting (\ref{eq_dbdq}), (\ref{eq_dak1}) and (\ref{eq_vdpjv}) into (\ref{eq_ckq}) yields:
\begin{equation}
        \begin{aligned}
                &(\mathbf{C}^{{\mathbf{p}_i}}_{m,n})_k =  \begin{bmatrix}
                        (\mathbf{C}^{x_i}_{m,n})_k & (\mathbf{C}^{y_i}_{m,n})_k & (\mathbf{C}^{z_i}_{m,n})_k
                       \end{bmatrix}_{1\times3} \\
                &=\overbrace{\frac{(k-1)(\lambda_{n} - \lambda_{m})_{k-1}}
                {k(\lambda_{n} - \lambda_{m})_k} \cos\theta_m \cos\theta_n}^{\mathbf{W}^{\mathbf{p}_i}_{m,n}}
                (\mathbf{C}^{\mathbf{p}_i}_{m,n})_{k-1}   \\
                &-\frac{d_u} {k^2(\lambda_{n} - \lambda_{m})_k}
                (\cos\varphi_{m}\mathbf{v}_{n}^T + \cos\varphi_{n}\mathbf{v}_{m}^T)_k
        \end{aligned}
        \label{eq_a_ckck1}
\end{equation}
where $\theta_m$ denotes the angle between $\mathbf{v}_{m,k}$ and $\mathbf{v}_{m,k-1}$.

For $\mathbf{p}_k$, which is independent of $\mathbf{A}_{k-1}$, \eqref{eq_ckq} simplifies to:
\begin{equation}
        \begin{aligned}
                &(\mathbf{C}^{\mathbf{p}_k}_{m,n})_k = \begin{bmatrix}
                        (\mathbf{C}^{x_k}_{m,n})_k & (\mathbf{C}^{y_k}_{m,n})_k & (\mathbf{C}^{z_k}_{m,n})_k
                       \end{bmatrix}_{1\times3} \\
                &=\frac{k-1}{ k^2 (\lambda_{n} - \lambda_{m})_k} \mathbf{v}_{m,k}^T \frac{\partial \mathbf{D}}{\partial \mathbf{p}} \mathbf{v}_{n,k} \\
                &=\frac{d_u(k-1)}{ k^2 (\lambda_{n} - \lambda_{m})_k} (\cos\varphi_{m}\mathbf{v}_{n}^T + \cos\varphi_{n}\mathbf{v}_{m}^T)_k.
        \end{aligned}  
        \label{eq_a_dbdqjk}
\end{equation}

\bibliographystyle{IEEEtran}
\bibliography{IEEEabrv, paper}

\end{document}